\documentclass[12pt]{article}
\usepackage[a4paper, hmargin=2.5cm, vmargin=2.5cm]{geometry}
\usepackage{amsfonts,amsmath}
\usepackage[authoryear]{natbib} \bibpunct{(}{)}{,}{a}{,}{,}

\usepackage{amsfonts,amsmath, amsthm}
\usepackage{comment}
\usepackage{booktabs}
\usepackage{caption}
\usepackage{float}
\usepackage{bm}
\usepackage{graphicx}

\newcommand{\bY}{\mbox{\boldmath $Y$}}

\newcommand{\bI}{\mbox{\boldmath $I$}}
\newcommand{\bX}{\mbox{\boldmath $X$}}

\newcommand{\by}{\mbox{\boldmath $y$}}

\newcommand{\bx}{\mbox{\boldmath $x$}}

\newcommand{\bPsi}{\mbox{\boldmath $\Psi$}}
\newcommand{\btheta}{\mbox{\boldmath $\theta$}}

\newcommand{\bmu}{\mbox{\boldmath $\mu$}}
\newcommand{\bSigma}{\mbox{\boldmath $\Sigma$}}
\newcommand{\bomega}{\mbox{\boldmath $\omega$}}

\newcommand{\bLambda}{\mbox{\boldmath $\Lambda$}}
\newcommand{\bbeta}{\mbox{\boldmath $\beta$}}

\newcommand{\om}{\overline{m}}

\newcommand{\hR}{\hat{\mbox{$R$}}}

\newcommand{\bxi}{\mbox{\boldmath $\xi$}}

\newcommand{\hbbeta}{\hat{\mbox{\boldmath $\bbeta$}}}
\newcommand{\hbtheta}{\hat{\mbox{\boldmath $\theta$}}}

\newcommand{\hbPsi}{\hat{\mbox{\boldmath $\bPsi$}}}

\newcommand{\mat}[1]{\bm{#1}}

\newcommand{\vect}[1]{\bm{#1}}

\newtheorem{lemma}{Lemma}
\newtheorem{theorem}{Theorem}

\usepackage{color}

\begin{document}
\title{Semi-Supervised Learning of Classifiers
from a  Statistical Perspective: A Brief Review}
\author{Daniel Ahfock,  Geoffrey J. McLachlan$^{\star}$}
\date{}

\maketitle

\begin{flushleft}
School of Mathematics and Physics, University of Queensland,
St. Lucia 4072, Australia.\\
$^\star$ E-mail: g.mclachlan@uq.edu.au
\end{flushleft}

\begin{abstract}
There has been increasing attention to semi-supervised learning
(SSL) approaches in machine learning to forming a classifier
in situations where the training data for a classifier
consists of a  limited number of classified observations but
a much larger number of unclassified observations.
This is because the procurement of classified data
can be quite costly due to high acquisition costs
and subsequent financial, time, and ethical issues
that can arise in attempts to provide the
true class labels for the unclassified data
that have been acquired.
A review is provided here
of statistical SSL approaches to this problem, focussing on the recent
result that a classifier formed from a partially classified
sample can actually have smaller expected error rate than that if the sample
were completely classified.
This rather paradoxical outcome is able to be achieved
by introducing a framework with a  missingness mechanism for the missing labels
of the unclassified observations.
It is most relevant in commonly occurring situations in practice,
where the unclassified data occur primarily in regions
of relatively high entropy in the feature space thereby making it
difficult for their class labels to be easily obtained.
\end{abstract}



\section{Introduction}
\label{intro}

Due to the scarcity and often high acquisition
cost of labelled data, machine learning methods
that make effective use of large quantities
of unlabelled data are being increasingly used.
One such method is semi-supervised learning (SSL) where,
in addition to labelled data, possibly large numbers
of unlabelled observations are available at the time
of the construction of the classification rule (classifier)
to be used.
Not surprisingly, 
semi-supervised learning approaches have been gaining much attention
in both the application oriented and the theoretical machine
learning communities.

However, theoretical analysis of SSL has so far been scarce.
But recently, \citet{AM20}
provided an asymptotic basis on how to increase
in certain situations the accuracy of
the commonly used linear discriminant function
formed from a partially classified sample as in SSL.
The increase in accuracy
can be of sufficient magnitude for this SSL-based
classifier to have smaller error rate than that
if it were formed from a completely classified sample.
This apparent paradox can be explained by the fact that
by introducing the concept of missingness
for the unobserved (missing) labels,
the information in the missing labels
has the potential to compensate
for the loss of information
from not knowing the missing labels
to the extent where this rule can have lower error rate than that
if the sample were completely classified in situations
where the missingness-mechanism of the class labels 
is non-ignorable in the pioneering framework of \citet{r76}
for missingness in incomplete-data analysis.

In this paper, we are to focus on SSL from a statistical
perspective reviewing, in particular, recent results.
Also, we provide some new results
in Lemma 1 and  Theorems and 3 and 4
on the asymptotic relative efficiency
of the classifier formed ignoring the information that is available
in situations where the labels of the unclassified features
are missing at random but non-ignorable.

\section{Brief Overview of SSL Approaches}

\noindent
Methods for semi-supervised learning (SSL)
typically involve one or more of the
following assumptions. 

\begin{itemize}
\item[(i)] 
{\bf Smoothness assumption:} If two feature vectors 
$\by_1$ and $\by_2$
in a high-density region are close,
then so should be their corresponding outputs (class labels)
$z_1$ and $z_2$.

\item[(ii)] 
{\bf Cluster assumption:} If points are in the same cluster, 
they are likely to be of the same
class.
\item[(iii)]
{\bf  Low-density separation assumption:}
The decision boundary should lie in a low-density
region.
\item[(iv)]
{\bf  Manifold assumption:}
High-dimensional data lie (roughly) on a low-dimensional
manifold.
\end{itemize}

These modelling assumptions create a paradigm 
where unlabelled data can be useful for model estimation
\citep{csz10}.
There has been a recent revival of interest 
in semi-supervised learning in the machine learning community 
due to impressive empirical progress on benchmark image 
and text classification datasets; for example,
\citet{tv17},
\citet{la17},
\citet{mmk19},
\citet{berth19},
\citet{xdh19}, 
\cite{berth20},
\citet{sb20},
and
\citet{wwk21}.

From a statistical perspective, the theoretical understanding 
of these modern semi-supervised learning algorithms 
is still limited.
One of the most intuitive approaches to 
semi-supervised learning is self-training, 
an iterative method for learning with 
alternating steps between generating pseudo-labels 
for the unlabelled observations and 
then training a classifier using both the 
labelled and pseudo-labelled data. 
As the algorithm progresses both the classification rule 
and the pseudo-labelling rule are updated. 
The underlying intuition is very similar 
to the underlying principles of the
expectation--maximization (EM) algorithm
of \citet{dlr77}, 
in that one iterates between generating pseudo-labels 
to form the complete-data and then estimating a classification
model on the basis of the complete-data. 
In early approaches, the same model is used for the
classification rule and the generation of pseudo-labels. 
However, this has been found to be inefficient
in some cases \citep{csz10}.
Modern approaches often use separate models for the classifier
and the pseudo-labeller, 
although the training of each can be coupled 
\citep{cdr20b,tv17,wgsh20}. 
Self-training can also be used to check that estimated error
rates for a classification rule 
are compatible with certain targets, such as homogeneous false positive
rates across demographic groups. 
This is particularly relevant when the fairness 
or algorithmic bias is a concern \citep{jss20}.
Unlabelled data are also used to construct 
data-dependent regularisation terms. 
Two of the most important types are 
entropy regularisation and consistency regularisation. 
Entropy regularisation enforces the belief that the 
classifier should make confident predictions 
on the unlabelled data \citep{gb05}.
Entropy regularisation is connected 
to the low-density separation assumption, 
as models with the decision boundary 
in low-density regions will generate predictions
with smaller expected entropy compared to models 
which place the decision boundary through a 
high-density region. Consistency regularisation 
encourages the model to make similar predictions
for unlabelled observations that are close together 
in feature-space \citep{belk06}. 
This is closely connected to the semi-supervised 
smoothness assumption.

Data-augmentation is another important technique 
that is often used to improve data-dependent regularisation. 
Additional unlabelled observations are generated 
by perturbing the original set of unlabelled observations 
by adding random noise or transformations 
such as rotations or translations. Data augmentation is typically coupled 
with consistency regularisation 
so that similar predictions are encouraged on
the original instances and the augmented versions 
\citep{berth19,nab19,wwk21}.  
The combined use of small local alterations 
and more aggressive global changes has been found to be 
an effective strategy \citep{sb20}.
The generating process for the augmentations can be related to 
the manifold assumption, as ideally the 
transformed data-points help to learn the 
natural low-dimensional manifold more accurately. 
This is particularly relevant when working with 
images \citep{hmn20}. 
Data augmentation has been demonstrated 
to improve the effectiveness of consistency regularisation 
and entropy regularisation 
\citep{berth19,nab19,sb20,wwk21}.

There are many open research questions 
regarding semi-supervised learning.
A fundamental issue is identifying the mechanisms 
responsible for the success of semi-supervised learning.
For generative classifiers that model the joint distribution 
of the features and the labels \citep{M92}, 
it is possible to compute the Fisher information 
in unclassified observations and the benefits
of using unlabelled observations are well understood. 
For discriminative classifiers that model the
conditional distribution of the labels given the features
\citep{nj02}, 
quantifying the available information 
in the unlabelled data is a very difficult problem 
\citep{LW07,snz08,gb19}.
There are some theoretical results
that show unlabelled data can improve estimation efficiency 
\citep{scy08,yr20}.
However, they do not fully explain the success 
of data-dependent regularisation and data augmentation. 
Recent theoretical work has highlighted the ability of data 
augmentation to encode prior knowledge about 
the classification problem, by enforcing invariance patterns 
that should hold under the generative model 
\citep{cdl20a,ds02}. 
The exploitation of task invariances is a general strategy 
for variance reduction in machine learning 
\citep{ds02,tgrs08}.

\section{\bf Optimal classifier specified by the Bayes' rule}

\noindent
In the sequel, we focus on statistical approaches to SSL
using parametric approaches to the estimation of the (optimal) Bayes' rule
from a partially classified sample.

We let $\by$ be a $p$-dimensional vector of features 
on an entity to be assigned to one of $g$ predefined classes 
$C_1,\,\ldots,\,C_g$.  
The random variable $\bY$ corresponding to the 
realisation $\by$ is assumed to have density 
$f_i(\by;\bomega_i)$ known up to a vector $\bomega_i$
of unknown parameters in Class $C_i\, (i=1,\,\ldots,\,g)$.
The optimal (Bayes') 
rule of allocation $R(\by;\btheta)$ assigns an entity
with feature vector $\by$ to Class $C_k$ 
(that is, $R(\by;\btheta)=k)$ if
\begin{equation}
k=\arg\max_i\; \tau_i(\by;\btheta),
\label{eq:1}
\end{equation}
where
\begin{equation}
\tau_i(\by;\btheta)= {\pi_i f_i(\by;\bomega_i)}/
{\sum_{h=1}^g \pi_h f_h(\by;\bomega_h)}
\label{eq:2}
\end{equation}
is the posterior probability that the entity belongs 
to Class $C_i$ given $\bY=\by$ and $\pi_i$ is the 
prior probability that the entity belongs to 
$C_i\,(i=1,\,\dots,\,g)$;
$\btheta=(\pi_1,\,\ldots,\,\pi_{g-1},
\btheta_1^T,\,\ldots,\,\btheta_g^T)^T$
is the vector of unknown parameters. 
The superscript $T$
denotes vector transpose.

In order to estimate $\btheta$, it is customary in practice to have 
available a training sample of size $n$.  We let
$\bx_{\rm CC}=(\bx_1^T,\,\ldots,\,\bx_n^T)^T$
contain $n$ independent realisations of $X=(\bY^T, Z)^T$  
as the completely classified training data, where $Z$ denotes
the class membership of $\bY$, being  equal to $i$ if $\bY$ 
belongs to class $C_i\ (i=1,\,\ldots,\,g)$, 
and zero otherwise, and 
where $\bx_j=(\by_j^T, z_j)^T$.
For a partially classified training sample
$\bx_{\rm PC}$ in SSL, we introduce 
the missing-label indicator $m_j$ which equals 1 if $z_j$
is missing and 0 if it is available $(j=1,\,\ldots,\,n)$.
Thus $\bx_{\rm PC}$ consists of those observations $\bx_j$
in $\bx_{\rm CC}$ with $m_j=0$, but only the feature vector
$\by_j$ in $\bx_{\rm CC}$ 
if $m_j=1$\,(that is, the label $z_j$ is missing).

Considerable simplification is possible under the two-class
(homoscedastic) normal model,
\begin{equation}
\bY \mid Z=i\; \sim \,N(\bmu_i, \bSigma)\quad {\rm in}\;\; C_i\;\;\;
{\rm with\; prob.}\;\; \pi_i\quad (i=1,2).
\label{eq:3}
\end{equation}
Under (\ref{eq:3}), the Bayes' rule
reduces to depending on just the $(p+1)$-dimensional vector of
discriminant function coefficients $\bbeta=(\beta_0,\bbeta_1^T)^T$,
since $R(\by;\btheta)$ is 1 or 2, according as 
\begin{equation}
d(\by;\bbeta)= \beta_0 + \bbeta_1^T\by
\label{eq:4}
\end{equation}
is greater or less than zero, where
\begin{eqnarray}
\beta_0 &=&\log(\pi_1/\pi_2)
-{\textstyle\frac{1}{2}}(\bmu_1+\bmu_2)^T\bSigma^{-1}
(\bmu_1-\bmu_2)\label{eq:5},\\
\bbeta_1 &=& \bSigma^{-1} (\bmu_1-\bmu_2).
\label{eq:6}
\end{eqnarray}

We can reparameterize the two-class normal model 
(\ref{eq:3}) by taking
$\btheta=(\btheta_1^T,\bbeta^T)^T,$
where $\btheta_1$ contains the elements of $\bmu=\pi_1\bmu_1+\pi_2\bmu_2$
and the distinct elements of
$$\bLambda=\bSigma+\pi_1\pi_2(\bmu_1-\bmu_2)
(\bmu_1-\bmu_2)^T.$$
It has the convenient canonical form,
\begin{equation}
\bSigma=\bI_p,\;
\bmu_1=(\Delta,0,\,\ldots,\,0)^T,\; \bmu_2=(0,\,\ldots,\,0)^T,
\label{eq:6a}
\end{equation}
where $\bI_p$ is the $p\times p$ identity matrix and
$$\Delta^2=(\bmu_1-\bmu_2)^T\bSigma^{-1}\bmu_1-\bmu_2)$$
is the squared Mahalanobis distance between the two classes.

\section{History of SSL in Statistics}

\noindent
In the statistical community, the so-called self-training approach 
to SSL in a sense was suggested by C.A.B.\ Smith (FRS), 
the distinguished British statistician and geneticist, 
who held the Weldon Chair of Biometry at University College London. 
In his discussion
of the paper titled 
\textit{Allocation Rules and Their Error Rates}
read to the Royal Statistical Society by \citet{h66}, 
\citet{cabs66} suggested that in the case of a completely unclassified 
sample which exhibits bimodality on some feature, 
a classifier be formed from the unclassified observations 
on the feature as follows: 
``One then arbitrarily divides them at the antimode, ....
On the basis of this division, we calculate a suitable allocation rule; 
and, by using this allocation rule, get
an improved division, and so on. 
As far as I know, there is no theoretical research into 
the effect of `lifting oneself by one's own bootstraps' in this way."

It subsequently led \citet{M75} to consider 
this iterative approach as suggested by \citet{cabs66}
on a theoretical basis. 
The parameters are estimated iteratively 
by treating the labels of the unclassified
features as unknown parameters to be estimated along with 
the parameters of the allocation rule.
That is, it uses the so-called classification
maximum likelihood (CML) approach as considered 
by \citet{hr68}, among others; 
see Section 1.12 of \citet{MB88}.
The CML approach gives an inconsistent estimate of
$\btheta$ except in special cases like
$\pi_1=\pi_2$.
The CML approach can be viewed as
applying the subsequent EM algorithm of \citet{dlr77}
with the following modification \citep{M82}.
Namely, the E-step is executed using outright (hard) 
rather than fractional (soft) assignment of each unclassified
feature to a component of the mixture as with 
the standard application of the EM algorithm.

In order to make the problem analytically tractable 
for the calculation of the expected error rate of the
estimated Bayes' rule, \citet{M75} assumed that 
there were also a limited number $n_{ic}$ of classified
features available from $C_i$ in addition to the 
number of $n_u = n-n_c$ unclassified features, 
where $n$ denotes the total size of the now partially 
classified sample and $n_c = n_{1c} + n_{2c}$.
We let $\hbbeta_{\rm PC}^{(k)}$ 
denote the estimator after the $k$th iteration 
of the vector $\bbeta=(\beta_0,\bbeta_1^T)^T$
of discriminant function coefficients obtained 
by the CML approach applied to the 
partially classified sample $\bx_{\rm PC}$. 
The estimated Bayes' rule using $\hbbeta_{\rm PC}^{(k)}$
for $\bbeta$ in the Bayes' rule $R(\by; \bbeta)$ 
is denoted by $\hR_{\rm PC}^{(k)}$.

\cite{M75} showed in the case 
of known equal prior probabilities 
that the overall expected error rate of this classifier
$\hR_{\rm PC}^{(k)}$
after the $k$th iteration
is given, as $n_u \rightarrow \infty$, by
\begin{equation}
\Phi(-{\textstyle\frac{1}{2}}\Delta)+
\{\phi({\textstyle\frac{1}{2}}\Delta)/4\}\,a_1^{(k)} +O(n_c^{-2}),
\label{eq:8}
\end{equation}
where
\begin{eqnarray*}
a_1^{(k)}&=& h_1^{2k}\frac{\Delta}{4} +h_2^{2k}\frac{p-1}{\Delta}
(\frac{1}{n_{1c}}+\frac{1}{n_{2c}})+h_2^{2k}\frac{(p-1)\Delta}{n_c-2},\\
\\
h_1&=&\phi({\textstyle\frac{1}{2}})
[4\phi({\textstyle\frac{1}{2}})+\Delta\{1-2\Phi(-{\textstyle\frac{1}{2}})],\\
h_2&=&\{\phi({\textstyle\frac{1}{2}})\}^2\,(4+\Delta^2)/h_1,
\end{eqnarray*}
where
$\phi(\cdot)$ and $\Phi(\cdot)$ denote the standard normal 
density and (cumulative) distribution, respectively, function.

As it can be shown that both $|h_1|$ and $|h_2|$ are always less than one, it 
follows from (\ref{eq:8}) that the expected error rate of 
$\hR_{\rm PC}^{(k)}$ decreases after each iteration 
and converges to the optimal error rate
$\Phi(-{\textstyle\frac{1}{2}}\Delta)$,
as $k \rightarrow \infty$.

\citet{M77} considered the above approach to estimating the 
linear discriminant function in the case where the number of unclassified
feature vectors was small rather than very large. 
The idea was to weight the estimates
of the class means based on the unclassified data with those based
on the classified data.
That is, the class estimates based on the classified data
were distinguished in importance from those based
on the unclassified data by the introduction of a 
weighting factor.
But it is preferable to weight each feature vector 
individually as to whether it is unclassified or not,
which is able to be  achieved 
with the application of the  EM algorithm which appeared
also in 1977.

More recently, a process called termed fractional
supervised classification (FSC) was proposed whereby 
the likelihoods $L_{\rm C}$ and $L_{\rm UC}$
were weighted by a single weighting factor
\citep{vm15,gm19}.
The reader is referred to \citet{MA21}
for some simulations on the FSC approach where the estimated
rule is applied to data subsequent to the partially
classified training sample.

\section{Estimation of the Bayes' Classifier}

\noindent
The construction of the parametric version
of the optimal (Bayes') classifier from partially classified data
can be undertaken by maximum likelihood (ML)
estimation of $\btheta$ implemented
via the expectation--maximization (EM) algorithm of
\citet{dlr77}; see also \citet{MK08}.
We let
\begin{eqnarray}
\log L_{\rm C}(\btheta)&=& 
\sum_{j=1}^n(1-m_j)\,\sum_{i=1}^g z_{ij}
\log\{\pi_i f_i(\by_j;\bomega_i)\}
\label{eq:9},\\
\log L_{\rm UC}(\btheta)&=&  
\sum_{j=1}^n m_j \log \sum_{i=1}^g \pi_i 
f_i(\by_j;\bomega_i),\label{eq:10}\\
\log L_{\rm PC}^{(\rm ig)}(\btheta)&=&\log L_{\rm C}(\btheta)
+\log L_{\rm UC}(\btheta),
\label{eq:11}
\end{eqnarray}
where in (\ref{eq:9}), $z_{ij}=1$ if $z_j=i$ and is zero otherwise.

In situations where one proceeds by ignoring
the ``missingness'' of the class labels,
$L_{\rm C}(\btheta)$ and $L_{\rm UC}(\btheta)$
denote the likelihood function formed from the classified data
and the unclassified data, respectively, and
$L_{\rm PC}^{(\rm ig)}(\btheta)$ is the likelihood function
formed from the partially classified sample $\bx_{\rm PC}$,
ignoring the missing-data mechanism for the labels.
The log of the likelihood $L_{\rm CC}(\btheta)$ for the
completely classified sample $\bx_{\rm CC}$ is given by (\ref{eq:9})
with all $m_j=0$.

In later sections, we are to consider situations 
where there is a link between
the pattern of missing labels and class uncertainty. 
In such situations, the likelihood function (\ref{eq:11}) 
is referred to as the likelihood that ignores 
the missing-label mechanism.

We let $\hbtheta_{\rm CC}$ and $\hbtheta_{\rm PC}^{(\rm ig)}$
be the estimate of $\btheta$ formed by consideration 
of $L_{\rm CC}(\btheta)$ and $L_{\rm PC}^{(\rm ig)}(\btheta)$,
respectively.
Also, we let
$R(\by;\hbtheta_{\rm CC})$ and  $R(\by;\hbtheta_{\rm PC}^{(\rm ig)})$ 
denote the estimated Bayes' rule obtained by plugging in
$\hbtheta_{\rm CC}$ and  $\hbtheta_{\rm PC}^{(\rm ig)}$, 
respectively, for $\btheta$ in $R(\by;\btheta)$.
The overall conditional error rate of the rule
$R(\by;\hbtheta_{\rm CC})$ is defined by
\begin{equation}
{\rm err}(\hbtheta_{\rm CC};\btheta)
= 1-\sum_{i=1}^g \pi_i\, {\rm pr}\{ R(\bY; \hbtheta_{CC}) = i
\mid \hbtheta_{CC}, Z=i\}.
\label{eq:12}
\end{equation}
The corresponding conditional error rate 
${\rm err}(\hbtheta_{PC}^{(\rm ig)}; \btheta)$ 
of the rule $R(\by;\hbtheta_{\rm PC}^{(\rm ig)})$
is defined by replacing $\hbtheta_{CC}$ with
$\hbtheta_{PC}^{(\rm ig)}$ in (\ref{eq:12}).
The optimal error rate ${\rm err}(\btheta)$
is defined by replacing  $\hbtheta_{\rm CC}$
by $\btheta$ in (\ref{eq:12}).

\section{ARE of $\hR_{\rm PC}^{(\rm ig)}$ in case of MCAR labels}

\noindent
The concepts of {\it missing at random\/} (MAR) 
and its more restrictive version 
{\it missing completely at random\/} (MCAR) 
are key concepts in the missing-data framework of 
\citet{r76} for missingness in incomplete-data analysis.
The reader is referred to the clarification note of 
of \citet{mr15}
for precise definitions of
MAR and MCAR and their corresponding more restrictive versions,
{\it missing always at random\/}
and {\it missing always completely at random\/},
respectively. 
Their paper also gives the weakest sufficient conditions
under which ignoring the missingness mechanism that creates the missing data 
always leads to appropriate inferences about $\btheta$ 
under frequentist, direct likelihood, or Bayesian modes of inferences.

In this section, it is assumed that the unclassified feature
vectors have labels that are MCAR.
The condition of the
missing data to be {\it missing (always) completely at random\/}
when coupled with exchangeability is a statement about
conditional independence in that it implies that
the vector of missing-label indicators $M_j$
is independent of the data
\citep{mr15}.
Here this implies that 
$M_j$ is independent of $\bY_j$ and $Z_j\,(j=1,\,\dots,\,n)$.

The relative value of the partially classified sample 
to the completely classified sample can be measured 
by comparing the expected excess error rate of a classifier 
trained on each of the respective samples $\bx_{\rm PC}$
and $\bx_{\rm CC}$
\citep{GM78,ON78,cc96}.

As seen in Section 3,
in the case of the two-class homoscedastic normal model 
(\ref{eq:3}), the Bayes' rule depends only on the vector
of discriminant coefficients $\bbeta$
as defined by (\ref{eq:5}) and (\ref{eq:6}).
We let $\hR_{\rm PC}^{(\rm ig)}=R(\by;\hbbeta_{\rm PC}^{(\rm ig)})$
and 
$\hR_{\rm CC}=R(\by;\hbbeta_{\rm CC})$
be the estimated Bayes' rules obtained by plugging the estimates
$\hbbeta_{\rm PC}^{(\rm ig)}$
and $\hbbeta_{\rm CC}$,
respectively, for $\bbeta$ into (\ref{eq:4}).

The relative efficiency of 
$\hR_{\rm PC}^{(\rm ig)}$ based on the partially classified sample
$\bx_{\rm PC}$ compared to the CSL rule 
$\hR_{\rm CC}$ 
based on the completely classified sample $\bx_{\rm CC}$
is defined as
\begin{equation}
{\rm RE}(\hR_{\rm PC}^{(\rm ig)})
=\frac{E\{{\rm err}(\hbbeta_{\rm CC};\bbeta)\} -{\rm err}(\bbeta)}
{E\{{\rm err}(\hbbeta_{\rm PC}^{(\rm ig)};\bbeta)\}-{\rm err}(\bbeta)}.
\label{eq:13}
\end{equation}

The asymptotic relative efficiency (ARE) of the rule
$\hR_{\rm PC}^{(\rm full)}$ compared to the rule 
$\hR_{\rm CC}$ 
is defined as
\begin{equation}
{\rm ARE}(\hR_{\rm PC}^{(\rm ig)})
=\frac{AE\{{\rm err}(\hbbeta_{\rm CC};\bbeta)\} -{\rm err}(\bbeta)}
{AE\{{\rm err}(\hbbeta_{\rm PC}^{(\rm ig)};\bbeta)\}-{\rm err}(\bbeta)},
\label{eq:14}
\end{equation}
where $AE\{{\rm err}(\hbbeta_{\rm PC};\bbeta)\}$ and
$AE\{{\rm err}(\hbbeta_{\rm CC};\bbeta)\}$
denote the expansion of the expected error rate
$E\{{\rm err}(\hbbeta_{\rm PC};\bbeta)\}$ and of
$E\{{\rm err}(\hbbeta_{\rm CC};\bbeta)\}$, respectively,
up to and including terms of the first order with respect to
$1/n.$

Under the assumption that 
the class labels are 
{\it missing always completely at random\/}
(that is, the missingness of the labels does not depend on the data),
\citet{GM78} derived the 
ARE of $\hR_{\rm PC}^{(\rm ig)}$
compared to $\hR_{\rm CC}$
in the case of a completely unclassified sample $(\gamma = 1)$ 
for univariate features $(p = 1)$.
Their results are listed in Table 1 
for $\Delta = 1, 2, 3,$ and 4. 

\begin{table}[h]
\centering
\captionsetup{width=0.4\textwidth}
\captionsetup{justification=centering}
\caption{Asymptotic relative efficiency of $\hR_{\rm PC}^{(\rm ig)}$
compared to $\hR_{\rm CC}$}
\label{tab:are}
\begin{tabular}{@{}lllll@{}}
\toprule
$\pi_{1}$ & $\Delta=1$ & $\Delta=2$ & $\Delta=3$ & $\Delta=4$ \\ \midrule
0.1       & 0.0036       & 0.0591       & 0.2540      & 0.5585      \\
0.2       & 0.0025       & 0.0668       & 0.2972      & 0.6068      \\
0.3       & 0.0027       & 0.0800       & 0.3289      &0.6352      \\
0.4       & 0.0038       & 0.0941       & 0.3509      & 0.6522      \\
0.5       & 0.0051       & 0.1008      & 0.3592      & 0.6580      \\ \bottomrule
\end{tabular}
\end{table}


\citet{ON78} made use of a result of \citet{ef75}
on the asymptotic relative efficiency of logistic regression
to extend the results of \citet{GM78} 
to multivariate features and for arbitrary $\gamma$.
His results showed that this ARE was not sensitive to the
values of $p$ and does not vary with $p$ for equal class
prior probabilities. 
It can be seen from Table \ref{tab:are}
that the ARE of $\hR_{\rm PC}^{(\rm ig)}$
for a totally unclassified
sample is low, particularly for classes weakly separated as 
represented by $\Delta= 1$ in Table~1.

In order to obtain the ARE of $\hR_{\rm PC}^{(\rm ig)}$
compared to $\hR_{\rm CC}$,
\citet{ON78} derived the information matrix
$\bI_{\rm PC}(\bbeta)$
about the vector $\bbeta$ of discriminant function coefficients 
in the context of the two-class normal discrimination problem
with parameter vector $\btheta$.
It was assumed that the missingness of the labels 
did not depend on the data.
Using the likelihood 
$L_{\rm PC}^{(\rm ig)}(\btheta)$ that ignores 
the mechanism for the missing-label indicators,
he showed that $\bI_{\rm PC}(\bbeta)$ 
can be decomposed as
\begin{equation}
\bI_{\rm PC}(\bbeta)=\bI_{\rm CC}(\bbeta) 
-\om\bI_{\rm CC}^{(\rm lr)}(\bbeta),
\label{eq:15}
\end{equation}
where $\bI_{\rm CC}(\bbeta)$ is the information about $\bbeta$ 
in a completely classified sample $\bx_{\rm CC}, 
\bI_{\rm CC}^{\rm(lr)}(\bbeta)$
is the conditional information about $\bbeta$ under 
the logistic regression model
for the distribution of the class labels given
the features in $\bx_{\rm CC}$,
and $\om=\sum_{j=1}^n m_j/n$
is the proportion of unclassified features 
in the partially classified sample $\bx_{\rm PC}$.
It can be seen from \eqref{eq:15} that the 
loss of information due to the sample being partially 
classified is equal to $\om\bI_{\rm CC}^{(\rm lr)}(\bbeta)$.
The consequent decrease in the efficiency in estimating 
the Bayes' rule can be considerable 
as illustrated above in Table \ref{tab:are}.

In other work on the ARE of $R(\by;\hbbeta_{\rm PC}^{(\rm ig)})$
compared to $R(\by;\hbbeta_{\rm CC})$,
\citet{MG89}
and
\citet{MS95}
evaluated it where the unclassified
univariate features had labels
that were not MCAR but were
{\it missing (always) at random\/} (MAR)
due to truncation of the features.

In that context, $L_{\rm PC}^{(\rm ig)}(\btheta)$ is
still the appropriate likelihood for the estimation of $\btheta$ 
as the probability that an entity with feature vector $\by_j$
has a missing label 
depends only on $\by_j$;
that is, the missingness is ignorable.
The full likelihood
$L_{\rm PC}^{(\rm full)}(\bPsi)$ actually then
reduces to $L_{\rm PC}^{(\rm ig)}(\btheta)$. 
Note that the Fisher information is affected 
by ignoring the missingness 
in its calculation since the
distribution of $L_{\rm PC}^{(\rm ig)}(\btheta)$
depends on the 
missingness mechanism
\citep{mr15}.
However, it can be estimated by the observed information matrix 
given by the negative of the Hessian of
$\log L_{\rm PC}^{(\rm ig)}(\btheta)$.
In the past attention has tended to focus more on when the 
missingness can be ignored in forming the likelihood
rather than forming a full likelihood to exploit
the additional information by formulating
a model to describe the mechanism underlying
the missing data \citep{hs09}.

\section{Illustration of missing labels with high entropy}

\noindent
In many practical applications cluster labels 
will be assigned by experts. 
Manual annotation of the dataset can induce 
a systematic missingness mechanism. 
From their examination of some
partially classified datasets,
\citet{AM19, AM20} conjectured
that in many situations  
the probability of a particular feature vector
being unlabelled is related to the difficulty of classifying the 
observation correctly. 
As an example, suppose medical professionals 
are asked to classify each image from a set of MRI scans 
into three classes, tumour present, no tumour present, or unknown. 
It would seem reasonable to expect that the unclassified observations 
would correspond to images that do not present clear evidence 
for the presence or absence of a tumour. 
In such situations 
the unlabelled images would fall in regions 
of the feature space where there is class overlap.  
This led \citet{AM19, AM20} to argue that in these situations, 
the unlabelled observations carry additional information 
that can be used to improve the efficiency of the parameter estimation. 
They suggested the the difficulty of classifying an observation 
could be quantified using the Shannon entropy 
of an entity with feature vector $\vect{y},$ 
which is defined by
\begin{equation}
    e(\vect{y}, \vect{\theta}) = - \sum_{i=1}^{g}\tau_{i}(\vect{y}; \vect{\theta}) \log \tau_{i}(\vect{y}; \vect{\theta}),
\label{eq:16}
\end{equation}
where $\tau_{i}(\vect{y}; \vect{\theta})$ 
is the posterior probability (\ref{eq:2}) that the entity belongs 
to the $i$th class given the feature vector $\vect{y}$. 
This measure can be used to define regions in the feature space
of high class uncertainty. 

We present below some results for three datasets that
were examined by \citet{AM19}
to demonstrate
that the missingness mechanism for the missing labels
can be related to the Shannon entropy. 
For visualisation we find it useful here to work 
with the negative log entropy.

\subsection{Flow cytometry dataset}
We first report results
for a flow cytometry dataset from \citet{agh13}
The dataset consists of fluorescence measurements 
on $n=11,792$ cells using $p=3$ markers. 
Cluster labels were assigned manually 
by domain experts using specialised software 
for the analysis of flow cytometry data. 
Labels were assigned using a combination 
of user-defined `gates' 
that partition feature space into groups. 
There were $n_{u}=333$ observations 
that were not assigned to a group at the end 
of the manual gating process. 
Figure \ref{fig:datasets} shows a bivariate plot of the dataset. 
Black squares denote unlabelled observations. 
Clusters are plotted using different colours and shapes. 
The unlabelled observations appear to be in areas 
where there is some overlap between clusters. 
Unlabelled observations appear to be concentrated 
around class decision boundaries, 
supporting the general idea that experts will hesitate 
to label observations that are difficult to classify. 

We fitted a skew $t$-mixture model \citep{LM14}
to estimate 
the classification entropy of each observation. 
Row (a) in Figure \ref{fig:data_missingness} 
compares the negative log entropy of the labelled 
and unlabelled observations. 
Panel (i) compares kernel density estimates, 
and Panel (ii) compares the empirical cumulative distribution 
functions of the estimated entropy distributions 
in the labelled and unlabelled groups. 
Panel (iii) shows a Nadaraya-Watson kernel estimate 
of the labelling probability. 
From (i) and (iii), 
we can see that the unlabelled observations 
typically have higher entropy than the labelled observations. 
The estimated missing label probability in (iii) 
appears to be a decreasing function of the negative log entropy.

\subsection{Cardiotocography dataset}
We now report results for a second dataset
that is a subset of data from \citet{abg00}
The full dataset consists of 23 features 
extracted from cardiotocograms on 2126 infants.  
A panel of three obstetricians used the cardiotocograms 
to assess fetal state. 
Observations were labelled as normal, pathological,  or suspect 
given the expert consensus. 
We take the suspect observations to be unlabelled. 
The cardiograms were also assigned a morphological 
pattern $(1,\,\ldots,\, 10)$ using automated methods. 
We restricted attention to the observations 
with morphological patterns $5,6,9,$ and $10$ 
as the majority of the unlabelled observations 
are in these groups. 
The subset we considered has $n=670$ observations, 
with $n_{c}=402$ labelled observations 
and $n_{u}=268$ unlabelled observations. 
We performed dimension reduction using 
principal components analysis prior to clustering. 
We worked with the first two principal component scores. 
Figure \ref{fig:datasets} (b) shows the data subset. 
Normal observations are plotted as blue circles, 
pathological observations are plotted as red triangles. 
Suspect observations are plotted as black squares. 
The bulk of the unlabelled observations 
are concentrated between the normal and pathological groups. 
Unlabelled observations appear to be in regions 
where there is class uncertainty. 

We fitted a two-component skew $t$-mixture model
\citep{LM14} 
to the observed dataset. 
We then estimated the entropy of each observation. 
Row (b) in Figure \ref{fig:data_missingness} 
compares the negative log entropy of the labelled 
and unlabelled observations. 
Panel (i) compares kernel density estimates 
and panel (ii) compares the empirical cumulative 
distribution functions. 
Panel (iii) shows a Nadaraya-Watson kernel estimate 
of the labelling probability. 
From (i) and (ii), we can see that the unlabelled 
observations typically have higher entropy 
than the labelled observations. 
The estimated missing label probability in (iii) 
shows a downward trend with respect to the negative log entropy. 

\subsection{Gastrointestinal dataset}
The third and final dataset to be reported on here
concerns a subset of data from  
\citet{mpa16}. 
The raw dataset consists of 700 features extracted 
from colonoscopic videos on patients 
with gastrointestinal lesions. 
There are $n=76$ records. 
A panel of seven doctors reviewed the videos 
and determined whether the lesions appeared benign 
or malignant. 
We formed a consensus labelling 
using the individual expert labels. 
The observations where six or more of the experts 
agreed were treated as labelled giving $n_{c}=53$. 
Observations where fewer than six experts agreed 
were treated as unlabelled giving $n_{u}=23$. 

\begin{figure}[H]
    \centering
    \includegraphics[width=\textwidth]{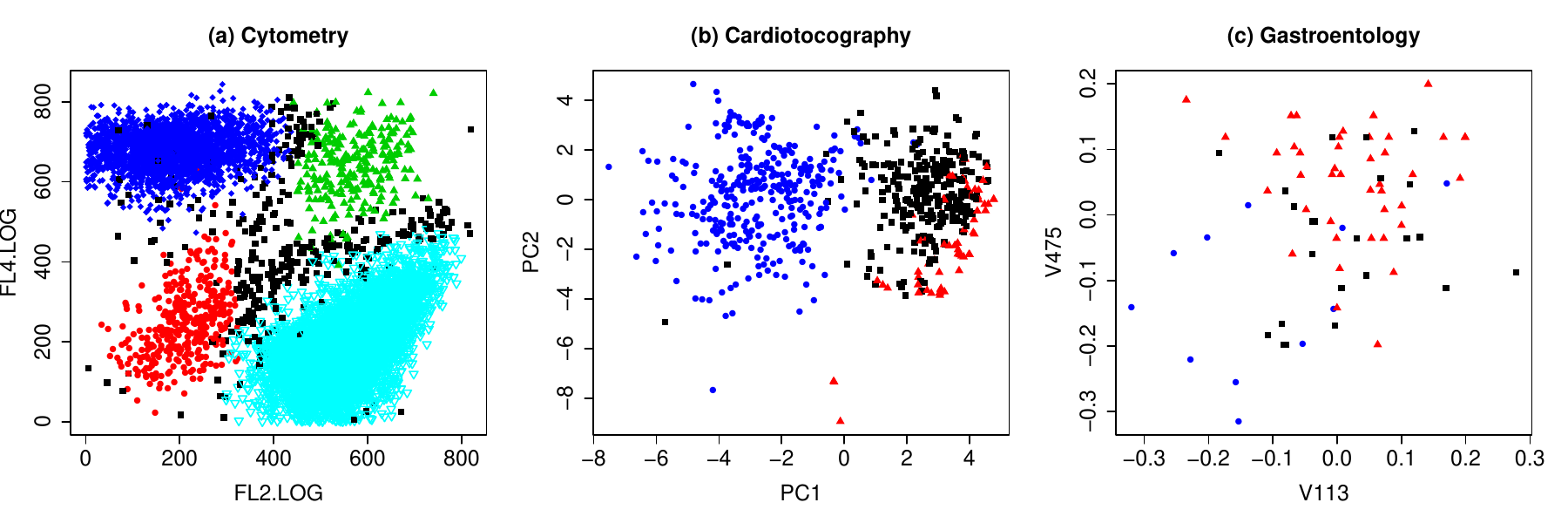}
    \caption{Partially classified datasets. Black squares denote unlabelled observations. Class membership for labelled observations is represented using different colours and shapes. (a) Flow cytometry dataset. (b) Cardiotocography dataset (c) Gastroentology dataset. }
    \label{fig:datasets}
\end{figure}

The dataset also includes a ground truth set of labels, 
obtained using additional histological measurements. 
The accuracy of the experts can be determined 
by comparing to the ground truth labels. 
To reduce the dimension of the dataset, 
we used sparse linear discriminant analysis 
\citep{chwe11} to select a subset 
of four features useful for class discrimination 
using the ground truth labels. 
These four variables were taken 
as the features for model based clustering. 
Figure \ref{fig:datasets} (c) shows 
a bi-variate plot of the data subset. 
Black squares denote unlabelled observations, 
red triangles denote benign observations, and 
blue circles denote malignant observations. 
It seems that the unlabelled observations 
are located in regions where there is group overlap. 
This dataset is smaller than the cytometry 
and cardiotocography datasets, 
so the pattern of missingness is less visually distinctive. 

\begin{figure}[H]
    \centering
    \includegraphics[width=\textwidth]{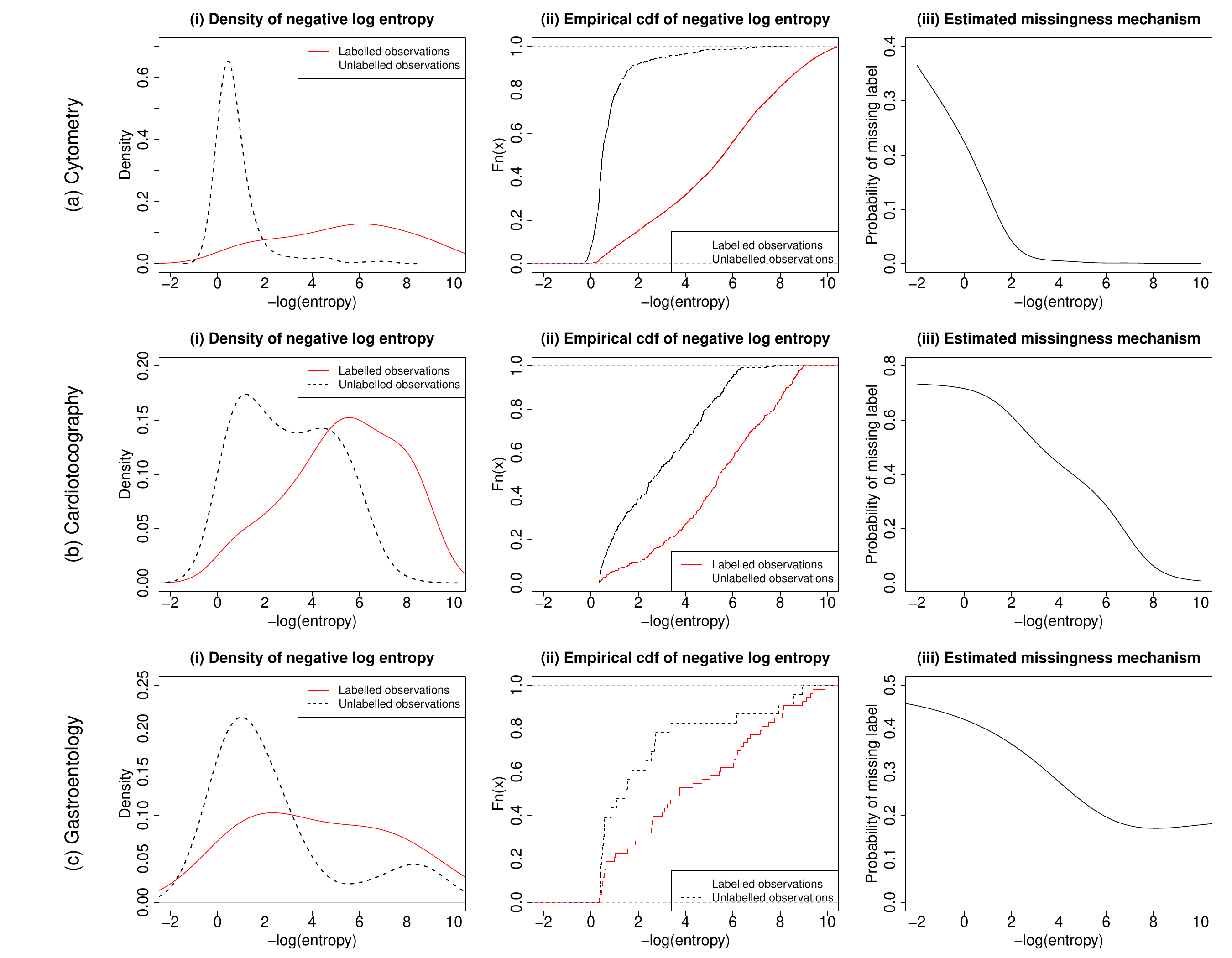}
    \caption{Analysis of missingness pattern in the partially classified datasets. For each dataset we fitted a model and estimated the entropy of each observation.  Row (a) shows results for the cytometry dataset, row (b) shows results for the cardiotocography dataset, and row(c) shows results for the gastroentology dataset. The first column shows density estimates of the negative log entropy. The second column compares the empirical cumulative distribution functions of the negative log entropy. Results for the unlabelled observations are shown as black dashed lines, and results for the labelled observations are shown as red solid lines. The third column shows a Nadaraya-Watson kernel estimate of the labelling probability as a function of the negative log entropy.}
    \label{fig:data_missingness}
\end{figure}

We fitted a two-component $t$-mixture model to the dataset. 
We then used the fitted model to estimate 
the entropy of each observation. 
Row (c) in Figure \ref{fig:data_missingness} 
compares the transformed entropy of the labelled 
and unlabelled observations. 
Panel (i) compares kernel density estimates 
and panel (ii) compares the empirical cumulative 
distribution functions. 
Panel (iii) shows a Nadaraya-Watson kernel estimate 
of the labelling probability. 
From (i) and (ii), we can see that the unlabelled 
observations typically have higher entropy 
than the labelled observations. 
There appears to be a relationship 
between the entropy and the estimated missing label 
probability in (iii).

\section{Modelling Missingness for Unobserved Class Labels}

\noindent
Following on from the exploratory examination
by \citet{AM19}
of partially classified data sets,
\citet{AM20} proposed to treat the labels of
the unclassified features as missing data and to introduce a
framework for their missingness as in 
\citet{r76} for missingness in incomplete-data analysis.
Within this framework, they postulated the dependence of the
conditional probability that a label is missing given the data
by the logistic model with covariate equal to an entropy-based
measure. 
This model was adopted bearing in mind that the
unclassified features of many datasets tend to fall in regions
of overlap of the classes in the feature space. 
This is not surprising as entities with features 
in such regions would tend to be representative 
of entities that would be difficult to classify correctly,
as illustrated for three datasets in the previous section. 
\citet{AM20} showed how
this dependency on the missingness pattern 
can be leveraged to provide additional information 
about the parameters in the optimal classifier 
specified by the Bayes' rule.

To this end,
they introduced the random variable
$M_j$ corresponding to the realized value $m_j$ for the
missing-label indicator for the feature vector $\by_j$.
The missing-data mechanism of Rubin (1976) is specified
in the present context by the conditional probabilities
\begin{eqnarray}
{\rm pr}\{M_j=m_j\mid \by_j, \,z_j\}
&=& {\rm pr}\{M_j=m_j\mid \by_j\} \label{eq:17}\\
&=& q(\by_j;\bPsi)\qquad \qquad \qquad (j=1,\,\ldots,\,n),
\label{eq:18}
\end{eqnarray}
where $\bPsi=(\btheta^T, \bxi^T)^T$
and where the parameter $\bxi=(\xi_0, \xi_1)^T$ 
is distinct from $\btheta$.

In the case where the probability (\ref{eq:18}) does not depend on 
$\btheta$ but only $\by_j$ and $\bxi$,
the missingness is ignorable as discussed in Section 6.
But now with this probability depending also on
$\btheta$, the missingness is non-ignorable.

On adopting the logistic function to model the dependence
of the probability (\ref{eq:18}) on the entropy of
the feature vector $\by_j$, or more precisely, the
negative log entropy,
we have  that
\begin{eqnarray} 
q(\by_j;\bPsi)&=& \frac{\exp\{\xi_0-\xi_1 \log e_j(\btheta)\}}
{1+\exp\{\xi_0-\xi_1 \log e_j(\btheta)\}},
\label{eq:19}
\end{eqnarray}
where the parameter $\bxi=(\xi_0, \xi_1)^T$ 
is distinct from $\btheta$.

The expected proportion $\gamma(\bPsi)$ of unclassified features
in a partially classified sample $\bx_{\rm PC}$ is given by
\begin{eqnarray}
\gamma(\bPsi)&=&\sum_{j=1}^n E(M_j)/n\nonumber\\
&=&E[{\rm pr}\{M_j=1\mid \bY_j\}]\nonumber\\
&=&E\{q(\bY;\bPsi)\}.
\label{eq:20}
\end{eqnarray}

To simplify the numerical computation in the particular case of
only $g=2$ classes with the 
two-class homoscedastic normal model (\ref{eq:3}),
\citet{AM20} replaced $\log e_j(\btheta)$
in (\ref{eq:19}) by minus the square of the discriminant
function $d_j=d(\by_j;\bbeta)$ as defined by (\ref{eq:4}) 
so that 
\begin{eqnarray} 
q(\by_j;\bPsi)&=& \frac{\exp\{\xi_0+\xi_1 d(\by_j;\bbeta)^2\}}
{1+\exp\{\xi_0+\xi_1 d(\by_j;\bbeta)^2\}}.
\label{eq:21}
\end{eqnarray}

In support of this approximation,
they noted that the negative log entropy
is a linear function of $d_j^2$ ignoring terms of order $O(d_j^{-4})$.

The probability of a missing label can also be expressed 
in terms of the value of the posterior probability of membership
of Class 1, 
$\tau_1(\vect{y}_{j}; \vect{\beta})$,
\begin{equation}
 {\rm pr} \{M_j=1\mid \by_j\} = q(\by_j;\bbeta,\bxi)= \frac{\exp\{\xi_0+\xi_1[\log (\tau_1(\vect{y}_{j}; \vect{\beta})/
\{1-\tau_1(\vect{y}_{j}; \vect{\beta})\})]^2\}}
{1+\exp\{\xi_0+\xi_1 [\log (\tau_1(\vect{y}_{j}; 
\vect{\beta})/\{1-\tau_1(\vect{y}_{j}; \vect{\beta})\})]^2\}}.
\label{eq:22}
\end{equation}

\begin{figure}
    \centering
    \includegraphics[width=\textwidth]{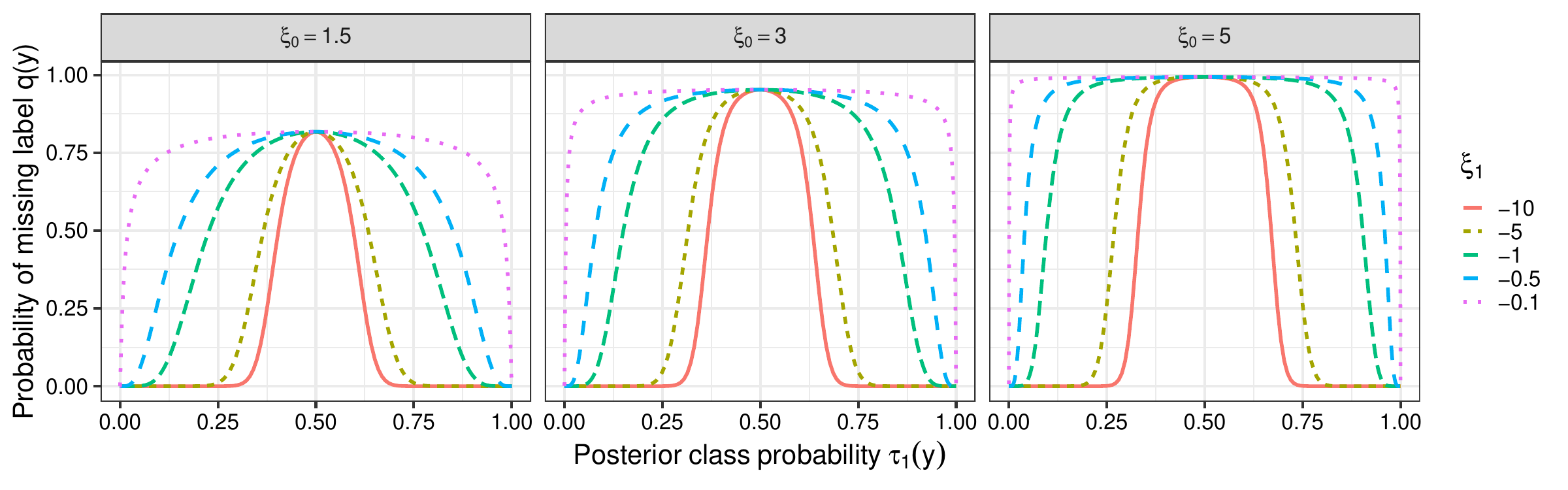}
    \caption{Probability of a missing label given 
different parameter values $\xi_{0}, \xi_{1}$ 
in the proposed missingness model. 
The probability of a missing label is greatest 
near the decision boundary where the posterior class probability 
is 0.5 and decreases as the classification difficulty decreases. 
The value of $\xi_{1}$ controls how rapidly the 
missingness probability decays moving away 
from the decision boundary. 
The value of $\xi_{0}$ has an important influence 
on the maximum probability of missingness 
that is attained on the decision boundary 
where the posterior class probability is 0.5.}
    \label{fig:missing_model}
\end{figure}

Figure \ref{fig:missing_model} 
plots the probability of missingness $ q(\by_j;\bbeta,\bxi)$ 
against the posterior class probability 
$\tau_1(\vect{y}_{j}; \vect{\beta})$ for the combinations 
of $\xi_{0}$ and $\xi_{1}$ that are used in Table 1. 
 The probability of a missing label has no further dependence 
on $\vect{\beta}$ 
given the posterior probability
$\tau_1(\by_j;\bbeta)$ of membership of Class 1.

\section{Full likelihood approach in case of non-ignorable MAR labels}

\noindent
We now consider in the case of non-ignorable missing class
labels the estimation of the Bayes' classifier 
as given by (\ref{eq:4}) under the two-class homoscedastic normal 
model (\ref{eq:3}).
We adopt the missing-label mechanism as proposed in
Section 8 whereby the probability
$q(\by_j;\bPsi)$ of the feature vector $\by_j$
having a missing label $(m_j=1)$,
is specified by the logistic model (\ref{eq:21}).
Thus the missingness is non-ignorable as this probability
now depends also
on $\btheta$, and hence, the parameters
of the Bayes' classifier.

The full likelihood function for $\bPsi$ 
is given by
\begin{equation}
\log L_{\rm PC}^{(\rm full)}(\bPsi)=
\log L_{\rm PC}^{(\rm ig)}(\btheta) +
\log L_{\rm PC}^{(\rm miss)}(\bPsi),
\label{eq:100}
\end{equation}
where
\begin{eqnarray}
\log L_{\rm PC}^{(\rm miss)}(\bPsi)&=&
\sum_{j=1}^n [(1-m_j)\log \{1- q(\by_j;\bPsi)\}
+ \, m_j \log q(\by_j;\bPsi)]
\label{eq:101}
\end{eqnarray}
is the log likelihood function for $\bPsi$
formed on the basis of the missing-label indicators
$m_j\,(j=1,\,\ldots,\,n)$
and $\log L_{\rm PC}^{(\rm ig)}(\btheta)$
is the log likelihood 
(\ref{eq:11}) formed ignoring the mechanism
for the missing class labels.

We let $\hbPsi_{\rm PC}^{(\rm full)}$
be the estimate of $\bPsi$ formed by consideration
of the full likelihood $L_{\rm PC}^{(\rm full)}(\bPsi)$ and
$\hR_{\rm PC}^{(\rm full)}=
R(\by;\hbbeta_{\rm PC}^{(\rm full)})$ be the estimated Bayes' rule
obtained by plugging in $\hbbeta_{\rm PC}^{(\rm full)}$ for
$\bbeta$ in the Bayes' rule $R(\by;\bbeta)$.

It was noted that there may be an identifiability issue
concerning $\bbeta$
and $\bxi$ if $\log L_{\rm PC}^{(\rm miss)}(\bbeta,\bxi)$
were to be used on its own for the estimation of $\bbeta$ and $\bxi$.
But as it is being combined with $\log L_{\rm PC}^{(\rm ig)}(\btheta)$
to form the full log likelihood $\log L_{\rm PC}^{(\rm full)}(\bPsi)$,
$\bbeta$ and $\bxi$ are each identifiable
with the use of the latter.

The asymptotic relative efficiency (ARE)
of the full likelihood rule $\hR_{\rm PC}^{(\rm full)}$
based on the partially classified sample $\bx_{\rm PC}$
compared to the rule $\hR_{\rm CC}$ based on the completely
classified sample $\bx_{\rm CC}$ is defined by
replacing $\hbbeta_{\rm PC}^{(\rm ig)}$ with 
$\hbbeta_{\rm PC}^{(\rm full)}$ in (\ref{eq:14}).
In order to derive the ARE of $\hR_{\rm PC}^{(\rm full)}$,
\citet{AM20}
derived the Fisher information about $\bbeta$ via the
full likelihood $L_{\rm PC}^{(\rm full)}(\bPsi)$.
It is given in the following theorem.

\noindent
\begin{theorem}
The Fisher information about $\bbeta$ 
in the partially classified sample $\bx_{\rm PC}$ 
via the full likelihood function $L_{\rm PC}^{(\rm full)}(\bPsi)$ 
can be decomposed as
\begin{equation}
\bI_{\rm PC}^{(\rm full)}(\bbeta)= \bI_{\rm CC}(\bbeta) 
-\gamma(\bPsi)\bI_{\rm CC}^{(\rm clr)}(\bbeta) + 
\bI_{\rm PC}^{(\rm miss)}(\bbeta),
\label{eq:100a}
\end{equation}
where
$\bI_{\rm CC}(\bbeta)$ is the information about $\bbeta$ 
in the completely classified sample $\bx_{\rm CC},
\bI_{\rm CC}^{(\rm clr)}(\bbeta)$ is the conditional information 
about $\bbeta$ under the logistic regression model
for the distribution of the class labels given
the features in $\bx_{\rm CC}$, and
$\bI_{\rm PC}^{(\rm miss)}(\bbeta)$ is the 
information about
$\bbeta$ in the missing-label indicators under the
assumed logistic model for their distribution 
given their associated features 
in the partially classified sample $\bx_{\rm PC}$.
\label{thm:inf}
\end{theorem}

It can be seen from (\ref{eq:100a}) that if 
\begin{equation}
\bI_{\rm PC}^{(\rm miss)}(\bbeta) > 
\gamma\bI_{\rm CC}^{(\rm clr)}(\bbeta),
\label{eq:101b}
\end{equation}
then there is actually an increase in the information about $\bbeta$ 
in the partially classified sample over
the information $\bI_{\rm CC}(\bbeta)$ about $\bbeta$
in the completely classified sample.
The inequality in (\ref{eq:101b}) is used 
in the sense that the left-hand side of the equation, 
minus the right, is positive definite.

\citet{AM20} showed that 
the inequality (\ref{eq:101b}) can hold
for various combinations of the parameters, which implies then
that $\hbbeta_{\rm PC}$
provides a more efficient estimator of the vector $\bbeta$ 
of discriminant function coefficients
than $\hbbeta_{\rm CC}$.
A consequence of this in such situations 
is that the asymptotic expected error rate
of the estimated Bayes' rule $\hR_{\rm PC}$ is actually
smaller than that of the rule
$\hR_{\rm CC}$ based on a completely classified sample.

The ARE of $\hR_{\rm PC}^{(\rm full)}$ 
compared to $\hR_{\rm CC}$ in the case of equal prior probabilities  
is given in Theorem \ref{thm:full}.
\begin{theorem}[\citet{AM20}] 
\label{thm:full}
Under the proposed missing-label model (\ref{eq:21}), 
the ARE of $\hR_{\rm PC}^{(\rm full)}$ 
compared to $\hR_{\rm CC}$ is given in the case of $\pi_{1}=\pi_{2}$ by
\begin{equation}
{\rm ARE}(\hR_{\rm PC}^{(\rm full)})=1 - (\Delta^2+4)(\gamma d_0-b_0)
\label{eq:23}
\end{equation}
for all $p$, where 
\begin{align}
   b_0&= \int_{-\infty}^{\infty}4\xi_1^2\Delta^2y_1^2q_1(y_1)
(1-q(y_1))f_{y_{1}}(y_1)dy_1,\nonumber \\
d_0&= \int_{-\infty}^{\infty}\tau_1(y_1)\tau_2(y_1)q_1(y_1)
\gamma^{-1}f_{y_1}(y_1)dy_1, \nonumber
\end{align}
and where 
\begin{align*}
\tau_1(y_1)&={\rm pr}\{Z=1\mid (\bY)_1=y_1\}\quad (i=1,2),\\
q_1(y_1;\Delta,\bxi)&={\rm pr}\{M=1\mid (\bY)_1=y_1\},\\
f_{y_1}(y_1;\Delta,\pi_1)&=\pi_1\phi(y_1;\Delta/2,1)
+(1-\pi_1)\phi(y_1;-\Delta/2,1).
\end{align*}
\end{theorem}
In Theorem \ref{thm:full}, the term $b_{0}$ is a function of 
$\mat{I}_{\rm PC}^{(\rm miss)}(\vect{\beta})$, 
and the term $d_{0}$ is related to the information loss 
due to the missing-labels $\gamma(\vect{\Psi})
\mat{I}_{\rm CC}^{(\rm clr)}(\vect{\beta})$. 

In the case of $\pi_1=\pi_2$,
Table 2 gives the ARE
of $\hR_{\rm PC}^{(\rm full)}$ compared to $\hR_{\rm CC}$
for various combinations of the parameters 
$\Delta, \xi_0$, and $\xi_1$,
the results applying for all values of $p$.
It can be seen for most of the combinations 
in Table 2 that 
the ARE of $\hR_{\rm PC}^{(\rm full)}$
is greater than one,
being appreciably greater than one 
for some combinations of the parameters.
For example, for $\Delta=1$ (representing classes close together)
or $\Delta=2$ (classes moderately separated), the ARE is not less than
14.8 for any combination with $\xi_0$ =3 or 5 and $\xi_1$=
$-5$ or $-10$, being as high as 40.4 for
$\Delta=1, \xi_0=5, \xi_1=-10$.
This shows
that the asymptotic expected excess error rate 
using the partially classified sample $\bx_{\rm PC}$
can be much lower than the corresponding excess rate 
using the completely classified sample $\bx_{\rm CC}$.

The general expression for the
ARE of $\hR_{\rm PC}^{(\rm full)}$ for $\pi_1\neq \pi_2$
is available in the supplementary material
of Ahfock and McLachlan (2020).
They noted 
that this ARE is not sensitive to the value $\pi_1$ 
in the range (0.2, 0.8), so that Theorem 2 can
provide useful guidelines for arbitrary prior probabilities.

\vfil
\eject

\begin{table}[h]
\centering
\captionsetup{width=0.4\textwidth}
\captionsetup{justification=centering}
\caption{\vspace{0.2cm} Asymptotic relative efficiency of $\hR_{\rm PC}^{(\rm full)}$
for $\pi_1=\pi_2$ (applicable for all $p$)}
\label{tab:theoretical}
\begin{tabular}{lrrrrrr}
\toprule
          &          & \multicolumn{5}{c}{$\xi_{1}$} \\ \cline{3-7} 
$\xi_{0}$ & $\Delta$ & -0.1  & -0.5  & -1  & -5 & -10 \\
\midrule
1.5      & 1       &0.2     & 1.5    & 3.6  & 15.0  & 23.1   \\
          & 2        & 0.8     & 3.1    & 4.7   & 10.3  & 14.4   \\
          & 3        & 1.6     & 2.9    & 3.6   & 6.6  &8.9  \\ \\
3      & 1        & 0.1    & 1.0    & 3.5   & 20.2  & 32.5   \\
          & 2        & 0.5     & 4.0    & 6.4   & 14.8  & 20.9   \\
          & 3        & 1.9     & 4.1   & 5.1   & 9.4  & 12.8  \\  \\
5      & 1        & 0.01     & 0.4    & 2.4   & 23.4  & 40.4   \\
          & 2        & 0.3     & 4.4    & 7.8   & 19.4  & 27.5   \\
          & 3        & 1.9     & 5.5    & 6.9   & 12.5  & 16.9  \\          
\bottomrule
\end{tabular}
\end{table}

\section{ARE of $\hR_{\rm PC}^{(\rm ig)}$ in the case of
non-ignorable MAR labels}

\noindent
Under the proposed missingness model \eqref{eq:21}, 
the labels are missing at random 
but the distinctness assumption (Little and Rubin, 2014) 
does not hold as the parameter vector $\btheta$ of the generative model 
enters the likelihood for the missing-label indicators. 
Even though the missingness mechanism is non-ignorable, 
the use of $L^{(\rm ig)}(\vect{\theta})$ still yields 
a consistent estimator, although it is no longer fully efficient. 
As $L^{(\rm ig)}(\vect{\theta})$ is a partial likelihood function, 
the asymptotic covariance matrix of the resulting estimator 
can still be determined by taking the inverse of the expectation 
of the negative of the Hessian matrix
(Little and Rubin, 2014; Little et al., 2017). 

In Lemma \ref{lem:ig}, we give 
an expression for 
$\bI_{\rm PC}^{(\rm ig)}(\bbeta)$ obtained on noting that
$$\mat{I}^{(\rm ig)}_{\rm PC}(\vect{\beta})=
\mat{I}_{\rm PC}^{(\rm full)}(\bbeta)
-\mat{I}_{\rm PC}^{(\rm miss)}(\vect{\beta}).$$
and using (\ref{eq:100a}) in Theorem \ref{thm:inf}.

\begin{lemma}
\label{lem:ig}
The information about $\bbeta$ 
in the partially classified sample $\bx_{\rm PC}$ 
using the likelihood that ignores the missingness 
mechanism $L_{\rm PC}^{(\rm ig)}(\bPsi)$ 
can be decomposed as
\begin{equation}
\bI_{\rm PC}^{(\rm ig)}(\bbeta)= \bI_{\rm CC}(\bbeta) 
-\gamma(\bPsi)\bI_{\rm CC}^{(\rm clr)}(\bbeta),
\label{eq:25}
\end{equation}
where
$\bI_{\rm CC}(\bbeta)$ 
is the information about $\bbeta$ 
in the completely classified sample $\bx_{\rm CC}$
and where
$\bI_{\rm CC}^{(\rm clr)}(\bbeta)$ is the conditional information 
about $\bbeta$ under the logistic regression model 
for the distribution of the class labels given 
the features in $\bx_{\rm CC}$.
\end{lemma}
\vspace{.5cm}

The ARE of $\hR_{\rm PC}^{(\rm ig)}$ 
compared to $\hR_{\rm CC}$ in the case of 
equal prior probabilities is given in Theorem \ref{thm:igcc}. 

\begin{theorem} 
Under the proposed missing-label model, 
the ARE of $\hR_{\rm PC}^{(\rm ig)}$ 
compared to $\hR_{\rm CC}$ is given in the case 
of $\pi_{1}=\pi_{2}$ by
\begin{equation}
{\rm ARE}(\hR_{\rm PC}^{(\rm ig)})=1 - (\Delta^2+4)\gamma d_0 \label{eq:26}
\end{equation}
for all $p$, where 
$d_{0}$ is defined as in Theorem \ref{thm:full}.
\label{thm:igcc}
\end{theorem}
The result follows from using Lemma 1 and Theorem \ref{thm:inf}.
The term $b_{0}=0$ in Theorem \ref{thm:full} is now zero since
$\mat{I}_{\text{PC}}^{(\rm miss)}(\vect{\beta})$
does not contribute to
$\mat{I}_{\rm{PC}}^{(\rm ig)}(\vect{\beta})$.

\vspace{.5cm}

The asymptotic relative efficiency of  $\hR_{\rm PC}^{(\rm ig)}$ 
compared to $\hR_{\rm PC}^{(\rm full)}$ can now
be obtained by combining the results in 
Theorems \ref{thm:full} 
and \ref{thm:igcc}.  It is given in Theorem \ref{thm:igpc}
in the case of equal prior probabilities.

\begin{theorem} 
Under the proposed missing-label model, 
the ARE of $\hR_{\rm PC}^{(\rm ig)}$ 
compared to $\hR_{\rm PC}^{(\rm full)}$ 
is given in the case of $\pi_{1}=\pi_{2}$ by
\begin{equation}
{\rm ARE}(\hR_{\rm PC}^{(\rm ig)} : \hR_{\rm PC}^{(\rm full)})=\dfrac{{\rm ARE}(\hR_{\rm PC}^{(\rm ig)})}{{\rm ARE}(\hR_{\rm PC}^{(\rm full)})},
\label{eq:27}
\end{equation}
where the numerator ${\rm ARE}(\hR_{\rm PC}^{(\rm ig)})$ 
is given by Theorem 3 and the denominator 
${\rm ARE}(\hR_{\rm PC}^{(\rm full)})$ is given by Theorem 2. 
\label{thm:igpc}
\end{theorem}

Table 3 reports the ARE of $\hR_{\rm PC}^{(\rm ig)}$
for different combinations of 
$\Delta, \xi_{0},$ and $\xi_{1}$. 
This ARE is very low for most of the combinations,
supporting the idea that incorporating the 
missingness mechanism can supply additional 
information and improve classification accuracy. 
Indeed, except for those combinations of the parameters 
with $\Delta<3$ and  $\xi_1 = -0.1$ in Table 3, the ARE is very low.  
It can be seen
for fixed values of $\Delta$ and $\xi_0$, 
the ARE decreases with $\xi_1$. 
Thus the case $\xi_1=-0.1$ corresponds to a situation where the
probability of a feature vector $\by_j$ having a missing label 
depends only weakly on the absolute value of the discriminant function
at this value of $\by_j$.

\begin{table}[h]
\centering
\captionsetup{width=0.475\textwidth}
\captionsetup{justification=centering}
\caption{\vspace{0.2cm} Asymptotic relative efficiency of $\hR_{\rm PC}^{(\rm ig)}$ compared to  $\hR_{\rm PC}^{(\rm full)}$
for $\pi_1=\pi_2$ (applicable for all $p$)}
\label{tab:theoreticala}
\begin{tabular}{lrrrrrr}
\toprule
          &          & \multicolumn{5}{c}{$\xi_{1}$} \\ \cline{3-7} 
$\xi_{0}$ & $\Delta$ & -0.1  & -0.5  & -1  & -5 & -10 \\
\midrule
  1.5 & 1 & 0.81 & 0.18 & 0.09 & 0.04 & 0.03 \\ 
   & 2 & 0.39 & 0.14 & 0.12 & 0.07 & 0.06 \\ 
   & 3 & 0.32 & 0.23 & 0.20 & 0.13 & 0.10 \\  \\
  3 & 1 & 0.78 & 0.09 & 0.04 & 0.02 & 0.02 \\ 
   & 2 & 0.29 & 0.07 & 0.06 & 0.04 & 0.04 \\ 
   & 3 & 0.21 & 0.13 & 0.12 & 0.08 & 0.07 \\  \\
  5 & 1 & 0.83 & 0.05 & 0.02 & 0.01 & 0.01 \\ 
   & 2 & 0.41 & 0.04 & 0.03 & 0.03 & 0.02 \\ 
   & 3 & 0.19 & 0.08 & 0.08 & 0.06 & 0.05 \\
\bottomrule
\end{tabular}
\end{table}

\subsection{Simulation}

We conducted a simulation to assess 
to what extent the asymptotic results
of the previous section apply in practice.
For each of the combinations of the parameters in Table 1, 
we generated $B = 1000$ samples of 
$\bX=(\bY^T \hspace{-0.25em},Z)^T$ to form the
completely classified sample $\bx_{\rm CC}$ and the
partially classified sample $\bx_{\rm PC}$.
On each replication,
the estimates 
$\hbbeta_{\rm PC}^{(\rm ig)}$ and $\hbbeta_{\rm PC}^{(\rm full)}$ 
were computed using a quasi-Newton algorithm, along with
the conditional error rates,
${\rm err}(\hbbeta_{\rm PC}^{(\rm ig)};\bbeta)$ 
and ${\rm err}(\hbbeta_{\rm PC}^{(\rm full)};\bbeta).$
We let ${\rm err}(\hbbeta_{\rm PC}^{({\rm ig}, b)};\bbeta)$ and 
${\rm err}(\hbbeta_{\rm PC}^{({\rm full}, b)};\bbeta)$ 
denote the conditional error rate 
 of $\hR_{\rm PC}^{(\rm ig)}$ and of
$\hR_{\rm PC}^{(\rm full)}$, respectively, 
on the $b$th replication.

The relative efficiency (RE) of $\hR_{\rm PC}^{(\rm ig)}$
compared to $\hR_{\rm PC}^{(\rm full)}$ was estimated by
\begin{equation}
\overline{{\rm RE}}(\hR_{\rm PC}^{(\rm ig)}:\hR_{\rm PC}^{(\rm full)};\bbeta)=
\frac{
B^{-1}{\sum_{b=1}^B \{{\rm err}(\hbbeta_{\rm PC}^{({\rm full}, b)};\bbeta)
-{\rm err}(\bbeta)}\}}
{B^{-1}{\sum_{b=1}^B 
\{{\rm err}(\hbbeta_{\rm PC}^{(\rm ig, b)};\bbeta)-{\rm err}(\bbeta)}\}}.
\label{eq:28}
\end{equation}

Table \ref{tab:n=500} reports the results for $n=500$, 
and Table \ref{tab:n=100} reports the results for $n=100$. 
As expected, the agreement with the theoretical 
asymptotic results is better for $n=500$ 
than with $n=100$. For $n=100$, the 
simulated relative efficiency is greater 
than one for the simulations with 
$\Delta=1,\xi_{1}=-0.1, \xi_{0}=1.5$ and 
$\Delta=1,\xi_{1}=-0.1, \xi_{0}=5$. 
The agreement between the simulation results 
and the theoretical values is 
good for $n=500$ and $\xi_{1} \ge 1$. 

\begin{table}[h]
\centering
\captionsetup{width=0.75\textwidth}
\captionsetup{justification=centering}
\caption{Simulated relative efficiency of $\hR_{\rm PC}^{(\rm ig)}$ compared to  $\hR_{\rm PC}^{(\rm full)}$
with $\pi_1=\pi_2$ for $n=500, p=1$ (Bootstrap standard errors
are in parentheses)}
\label{tab:n=500}
\begin{tabular}{lrrrrrr}
\toprule
          &          & \multicolumn{5}{c}{$\xi_{1}$} \\ \cline{3-7} 
$\xi_{0}$ & $\Delta$ & -0.1  & -0.5  & -1  & -5 & -10 \\
\midrule
 1.5 & 1 & 0.93 (0.026) & 0.18 (0.011) & 0.09 (0.005) & 0.04 (0.002) & 0.03 (0.002) \\ 
   & 2 & 0.39 (0.021) & 0.16 (0.009) & 0.12 (0.007) & 0.08 (0.005) & 0.06 (0.004) \\ 
   & 3 & 0.34 (0.018) & 0.22 (0.012) & 0.20 (0.011) & 0.14 (0.009) & 0.12 (0.008) \\ \\
  3 & 1 & 0.98 (0.041) & 0.09 (0.005) & 0.04 (0.002) & 0.02 (0.001) & 0.02 (0.001) \\ 
   & 2 & 0.31 (0.018) & 0.08 (0.005) & 0.06 (0.003) & 0.04 (0.003) & 0.03 (0.002) \\ 
   & 3 & 0.21 (0.011) & 0.13 (0.008) & 0.12 (0.008) & 0.09 (0.006) & 0.08 (0.005) \\ \\
  5 & 1 & 0.88 (0.028) & 0.14 (0.015) & 0.02 (0.001) & 0.01 (0.001) & 0.01 (0.001) \\ 
   & 2 & 0.44 (0.032) & 0.03 (0.002) & 0.03 (0.002) & 0.03 (0.002) & 0.02 (0.002) \\ 
   & 3 & 0.21 (0.013) & 0.08 (0.005) & 0.08 (0.005) & 0.07 (0.004) & 0.06 (0.004) \\   
\bottomrule
\end{tabular}
\end{table}

\begin{table}[h]
\centering
\captionsetup{width=0.75\textwidth}
\captionsetup{justification=centering}
\caption{Simulated relative efficiency of $\hR_{\rm PC}^{(\rm ig)}$ compared to  $\hR_{\rm PC}^{(\rm full)}$
with $\pi_1=\pi_2$ for $n=100, p=1$ (Bootstrap standard errors
are in parentheses)}
\label{tab:n=100}
\begin{tabular}{lrrrrrr}
\toprule
          &          & \multicolumn{5}{c}{$\xi_{1}$} \\ \cline{3-7} 
$\xi_{0}$ & $\Delta$ & -0.1  & -0.5  & -1  & -5 & -10 \\
\midrule
  1.5 & 1 & 1.12 (0.037) & 0.27 (0.029) & 0.10 (0.006) & 0.04 (0.003) & 0.03 (0.002) \\ 
   & 2 & 0.42 (0.020) & 0.16 (0.009) & 0.13 (0.008) & 0.10 (0.007) & 0.07 (0.004) \\ 
   & 3 & 0.33 (0.018) & 0.26 (0.015) & 0.25 (0.014) & 0.25 (0.015) & 0.29 (0.016) \\  \\
  3 & 1 & 1.05 (0.018) & 0.42 (0.028) & 0.09 (0.015) & 0.02 (0.002) & 0.02 (0.001) \\ 
   & 2 & 0.43 (0.046) & 0.07 (0.005) & 0.06 (0.003) & 0.06 (0.004) & 0.05 (0.003) \\  
   & 3 & 0.24 (0.015) & 0.16 (0.009) & 0.15 (0.009) & 0.17 (0.009) & 0.20 (0.012) \\ \\
  5 & 1 & 0.96 (0.019) & 0.79 (0.033) & 0.28 (0.024) & 0.01 (0.001) & 0.01 (0.001) \\ 
   & 2 & 0.92 (0.073) & 0.04 (0.003) & 0.04 (0.002) & 0.04 (0.003) & 0.04 (0.003) \\ 
   & 3 & 0.19 (0.012) & 0.10 (0.006) & 0.10 (0.007) & 0.15 (0.009) & 0.19 (0.012) \\ 
\bottomrule
\end{tabular}
\end{table}

\vfil
\eject

\section{Discussion}
\label{sec:con}

\noindent
The results presented here on a statistical SSL approach 
to classification
are quite encouraging in that they show there are gains
to be made in efficiency concerning
the estimation of the parameters in the Bayes' rule. 
Situations where these gains can be achieved are where the labels
of the unclassified features in the partially classified sample
are not missing completely at random (MCAR) but rather occur
according to a missing-label mechanism that is able to be modelled
appropriately.   
A common situation with labels that are missing at random (MAR)
but non-ignorable is one in which the
features that are difficult to classify correctly
are left unclassified.
In the theoretical and practical results presented in this review
the missingness mechanism has been represented 
via the use of the logistic
function to model the conditional probability that a label
is missing via its dependence on the log entropy of the feature vector.
The choice of the logistic function can be modified to handle
situations where the features with missing labels are confined 
to lying in only certain regions of high entropy.

As remarked in the introduction, the fact that the
full likelihood-based rule $\hR_{\rm PC}^{(\rm full)}$
can outperform the completely supervised learning (CSL)
rule $\hR_{\rm CC}$ is very surprising.
In situations where this can occur, that is, where the unclassified
features are those occurring in regions of high entropy
in the feature space, means that a large number of unclassified
observations are not needed to build a SSL classifier with
small error rate.

Related to the full likelihood SSL classifier having 
performance better or comparable to the CSL classifier
is that the SSL rule $\hR_{\rm PC}^{(\rm ig)}$
that ignores the missingness mechanism for the missing labels
performs very poorly relative to the full likelihood-based rule
$\hR_{\rm PC}^{(\rm full)}$. 
The asymptotic relative efficiency of $\hR_{\rm PC}^{(\rm ig)}$
compared to $\hR_{\rm PC}^{(\rm full)}$ has been derived
here using the results produced in  \citet{AM20}
during their derivation of the ARE of $\hR_{\rm PC}^{(\rm full)}$. 
The results so obtained for the ARE of 
$\hR_{\rm PC}^{(\rm ig)}$ 
show how it can be in situations where the missing-label
pattern is non-ignorable.

Simulation results undertaken by \citet{L66,L74}
and theoretical results provided initially by \citet{M72}
and, more recently, Cannings et al.\ (2020), 
have shown that the error rate of a classifier
can be adversely affected by misclassification
of its training data. 
Thus in situations where the SSL rule based on the
full likelihood $\hR_{\rm PC}^{(\rm full)}$ 
has better or comparable performance to the 
completely classified (CSL) rule $\hR_{\rm CC}$, 
one would expect it to do even better for a 
partially classified sample in which some of the
labelled features are misclassified.

In the results presented in this paper, 
the error rate of a classifier 
refers to its application to unclassified features not in
the partially classified sample. 
But in ongoing work, we also have been considering the error rates 
of the classifiers in their application to the unclassified features 
in the partially classified training sample.
Preliminary results show that the relative efficiency of the 
rules $\hR_{\rm PC}^{(\rm ig)}$ and 
$\hR_{\rm PC}^{(\rm full)}$ has not significantly changed. 
In the situations where the unclassified features in the sample are
those that are the most difficult to classify, it is to be expected
the actual sizes of the error rates of these rules
are greater than in their application to new randomly chosen features.
Also, the rule $\hR_{\rm PC}^{(\rm full)}$ will not be 
superior to the CSL rule $\hR_{\rm CC}$ 
since the latter is being applied to 
the unclassified features whose true labels were used 
in the construction of this CSL rule.
This provides the rule $\hR_{\rm CC}$ with
too great an advantage over the rule $\hR_{\rm PC}^{(\rm full)}$.

Attention has been confined here to the situation 
where the features are from a mixture of two classes
that have  multivariate normal distributions
with a common covariance matrix.
In this situation, it is possible to
investigate analytically the relative performance of the classifiers.
However, as the model of \citet{AM20} for the missingness 
of the class labels is defined in terms of the entropy of 
an observed feature, it is applicable to more than two 
classes with arbitrary distributions.
As part of our ongoing research 
on the full likelihood approach of \citet{AM20},
we intend to carry out simulations in such situations,
in particular, for
heteroscedastic multivariate normal distributions.

\vspace{.1cm}

\noindent
{\bf Acknowledgement}

\noindent
This research was funded in part by the Australian Government 
through the Australian Research Council (Project Numbers DP180101192 
and IC170100035).







\bibliographystyle{elsarticle-harv}
\bibliography{Refs}

\end{document}